\documentclass[runningheads]{llncs}
\usepackage{subfigure}

\usepackage[T1]{fontenc}
\usepackage{xcolor}

\usepackage{graphicx}
\begin{document}
\title{Predicting Brain Tumor Response using a Hybrid Deep Learning and Radiomics Approach}
\titlerunning{Hybrid Deep Learning and Radiomics Approach}
%
\author{%
Daniil Tikhonov\thanks{Equal contribution.} \and
Matheus Ferracciú Scatolin$^{\star}$ \and \\
Mohor Banerjee \and Qiankun Ji \and Ahmed Jaheen \and Damir Kim \and \\
Mostafa Salem \and Abdelrahman Elsayed \and Hu Wang \and 
Sarim Hashmi \and
Mohammad Yaqub
}

\authorrunning{D. Tikhonov, M. F. Scatolin et al.}
%

\institute{
Mohamed bin Zayed University of Artificial Intelligence, Abu Dhabi, UAE\\
\email{\{firstname.lastname\}@mbzuai.ac.ae} \\
}

\maketitle              
\begin{abstract}
Accurate evaluation of the response of glioblastoma to therapy is crucial for clinical decision-making and patient management. The Response Assessment in Neuro-Oncology (RANO) criteria provide a standardized framework to assess patients' clinical response, but their application can be complex and subject to observer variability. This paper presents an automated method for classifying the intervention response from longitudinal MRI scans, developed to predict tumor response during therapy as part of the BraTS 2025 challenge. We propose a novel hybrid framework that combines deep learning derived feature extraction and an extensive set of radiomics and clinically-chosen features. Our approach utilizes fine-tuned ResNet-18 model to extract features from 2D regions of interest across four MRI modalities. These deep features are then fused with a rich set of more than 4,800 radiomic and clinically-driven features, including 3D radiomics of tumor growth and shrinkage masks, volumetric changes relative to the nadir, and tumor centroid shift. Using the fused feature set, a CatBoost classifier achieves a mean ROC AUC of 0.81 and a Macro F1 score of 0.50 in the 4-class response prediction task (Complete Response, Partial Response, Stable Disease, Progressive Disease). Our results highlight that synergizing learned image representations with domain-targeted radiomic features provides a robust and effective solution for automated treatment response assessment in neuro-oncology.

\end{abstract}

\section{Introduction}
Glioblastoma is the most aggressive primary brain tumor in adults, characterized by rapid progression and a challenging prognosis. For patients undergoing treatment, typically involving surgical resection followed by chemoradiotherapy, longitudinal monitoring with Magnetic Resonance Imaging (MRI) is essential to evaluate treatment efficacy and guide subsequent clinical management \cite{Kickingereder2019}. Therefore, the accurate assessment of tumor changes between consecutive imaging time-points is essential to neuro-oncological practice.

To standardize this evaluation process, the Response Assessment in Neuro-Oncology (RANO) criteria were developed, providing a framework for classifying tumor response into four distinct categories: Complete Response (CR), Partial Response (PR), Stable Disease (SD), and Progressive Disease (PD) \cite{Wen2010}. While the RANO criteria have improved consistency, manual assessment remains a time-consuming task that is susceptible to inter-observer variability, particularly in complex cases. This has motivated a growing interest in the development of automated, quantitative methods to provide objective and reproducible response assessments.

The MICCAI Brain Tumor Segmentation (BraTS) challenge has been central in advancing the field of automated brain tumor analysis. The 2025 BraTS Challenge, specifically the sub-challenge on Brain Tumor Progression, addresses the critical gap of longitudinal response prediction by tasking participants with developing methods to classify tumor response according to the RANO criteria. The challenge provides the LUMIERE dataset for training, which includes longitudinal MRI scans with expert RANO annotations \cite{Suter2022}. 

Recent advancements in medical image analysis have been driven by two prominent methodologies: deep learning and radiomics. Deep learning models, particularly Convolutional Neural Networks (CNNs), excel at automatically learning hierarchical feature representations from imaging data. In parallel, radiomics focuses on extracting a large number of engineered, quantitative features from medical images, capturing information about tumor appearance, texture, and morphology. While both approaches have shown promise, methods that effectively integrate the strengths of both paradigms for longitudinal analysis remain an emerging area of research.

In this work, we present a novel hybrid approach developed for the BraTS challenge. Our contribution is a multi-faceted model for predicting RANO-defined tumor response, which includes the following key components:

\begin{itemize}
    \item Fusion of deep learning features with a comprehensive suite of radiomic, volumetric, and engineered features.
    \item Fine-tuning of four separate ResNet-18 architectures to extract modality-specific features from different MRI sequences.
    \item Integration of over 4,800 additional features capturing 2D and 3D tumor characteristics, including volumetric changes, growth and shrinkage patterns, and shifts in the tumor centroid.
    \item Demonstration that the combination of learned and engineered features results in a robust and accurate model for tumor response prediction.
\end{itemize}

\section{Methods}
Our methodology for predicting tumor response is structured as a multi-stage pipeline. We first normalize and preprocess the longitudinal MRI data. Then, we employ a hybrid model that extracts features using both deep learning-based encoders and an extensive set of handcrafted radiomic and engineered features. Finally, a gradient boosting classifier integrates these features to predict one of the four RANO classes.

\subsection{Dataset and Preprocessing}
The foundation of this work is the publicly available LUMIERE dataset \cite{Suter2022}, provided for the Task 11 of the BraTS 2025 challenge. This dataset contains 616 longitudinal MRI scans from 91 glioblastoma patients, with expert-annotated RANO ratings for consecutive scan pairs. Each scan includes up to four standard MRI sequences: T1-weighted (T1), post-contrast T1-weighted (T1c), T2-weighted (T2), and Fluid-Attenuated Inversion Recovery (FLAIR). We utilize the provided automatic tumor segmentations generated by the HD-GLIO-AUTO \cite{Kickingereder2019,Isensee2019,Isensee2019nnunet} and DeepBraTumIA \cite{Grun2023} tools.

\begin{figure}[!htb]
\begin{center}
\includegraphics[width=\columnwidth]{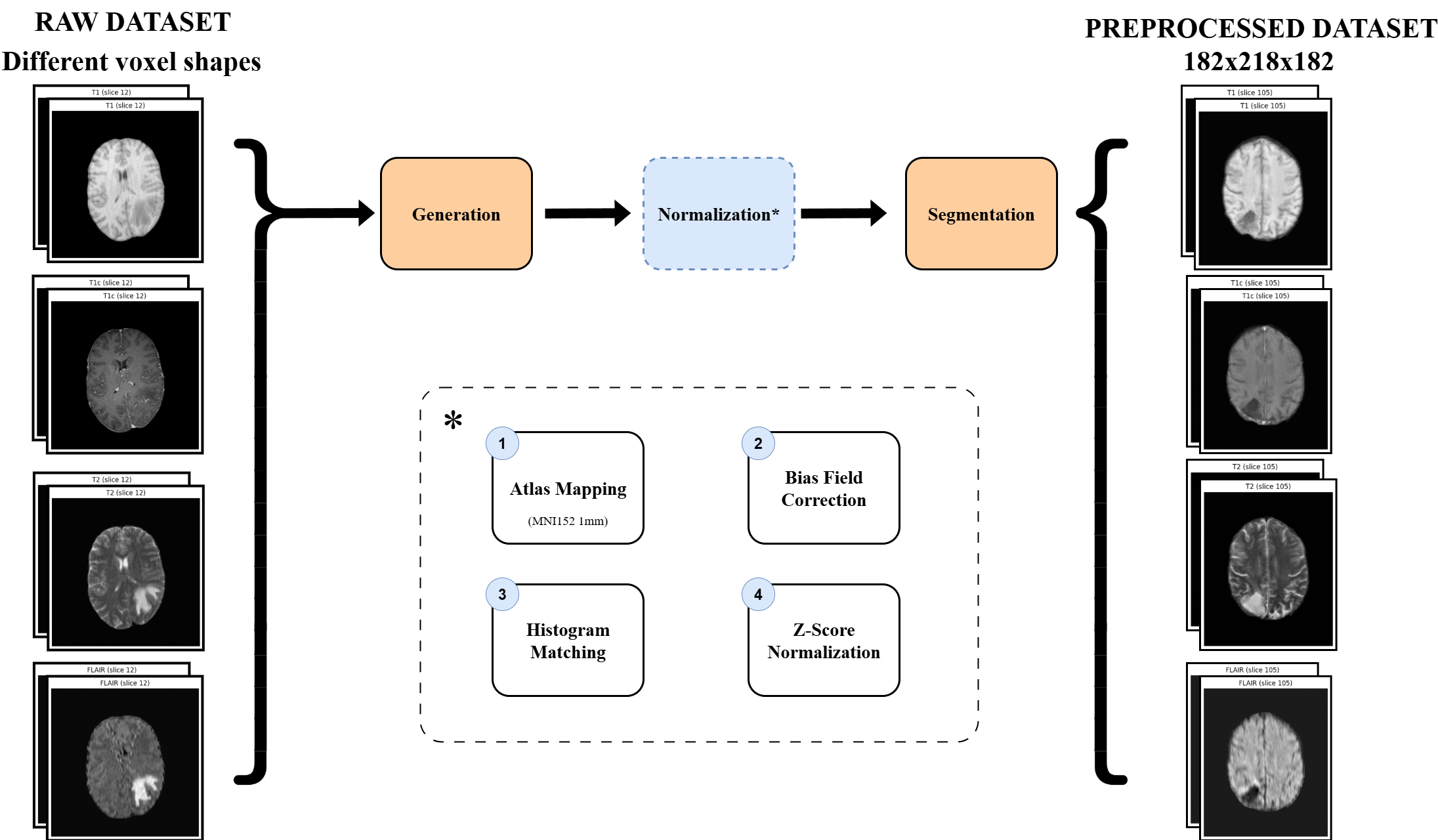}
\caption{Overview of the preprocessing pipeline applied to the raw MRI data. The central diagram represents the sequential transformation of the original scans, aiming to ensure consistency across all samples. The four surrounding boxes illustrate key normalization steps: (1) spatial registration to MNI152 space, (2) N4 bias field correction, (3) intensity normalization via histogram matching and (4) z-scoring. These steps standardize spatial and intensity properties of the data, enabling robust downstream analysis.}
\label{Fig1}
\end{center}
\end{figure}

A comprehensive preprocessing pipeline was applied to the raw MRI images (as illustrated in Figure~\ref{Fig1}) to ensure consistency and improve model performance. The key steps included filtering out samples with any missing MRI modalities to ensure a complete four-sequence set for every time-point, registration of all images to the MNI152 1mm standard space atlas \cite{Fonov2011}, application of the N4 bias field correction algorithm \cite{Tustison2010} to correct for intensity non-uniformities, histogram matching to standardize image intensities across all patient scans, and z-score normalization of the intensity values. All subsequent analysis was performed on these preprocessed images.

\subsection{Proposed Hybrid Model}
Our proposed model is a hybrid framework that leverages the distinct advantages of learned deep representations and domain-specific handcrafted features. For each baseline and follow-up case, feature sets from both categories are extracted and then concatenated before being passed to the final classifier.

\subsubsection{Deep Feature Extraction from MRI Scans.}
To capture complex patterns from the imaging data, we utilized deep learning encoders based on the ResNet-18 architecture \cite{He2016}. Instead of using the full 3D volume, we implemented a 2D Region of Interest (ROI) cropping strategy to focus the model on the most informative slice. For each 3D MRI scan, we first identified the single axial slice with the largest tumor segmentation area. A 128x128 pixel ROI was then extracted from this slice.

We fine-tuned four independent ResNet-18 models, one for each preprocessed MRI modalities (T1, T1c, T2, FLAIR). Each network was trained to act as an image encoder, processing the 2D ROI and outputting a compact feature vector representing the visual information. The image representations from all four modalities for both the baseline and follow-up scans are subsequently used as features in the final classification model.

To leverage transfer learning, we initialized each ResNet-18 with ImageNet-pretrained weights. For fine-tuning, we unfroze only the final residual block and the classification head, allowing the model to adapt high-level features to the domain-specific imaging data while preserving the robustness of the early convolutional layers. The input ROIs were resized to match the expected input size of the pretrained network.

Optimization was conducted using both SGD and AdamW optimizers across different configurations. A learning rate warm-up strategy was employed, ramping up to 1e-4 over the first 10 epochs, followed by cosine annealing to 1e-6 over 150 training epochs. We used weighted random sampling with inverse class frequencies to handle class imbalance in the dataset and employed the Focal Loss ($\gamma = 2.0$, and no class-specific alpha weighting). This loss function emphasizes harder examples during training, which improves robustness against class skew.

\subsubsection{Radiomic and Engineered Feature Extraction.}
To complement the learned representations from the ResNet encoders, we engineered a comprehensive set of handcrafted features designed to capture interpretable clinical and morphological characteristics of the tumor. These handcrafted features are grouped into three main categories.

First, we extracted an extensive set of 2D and 3D radiomic features using the Pyradiomics library \cite{VanGriethuysen2017}. To capture dynamic changes, features were calculated for both the baseline and follow-up scans, and across three distinct regions of interest: the \textit{original mask} of the entire tumor; a \textit{growth mask}, defined as the tumor region present in the follow-up scan but not in the baseline; and a \textit{shrinkage mask}, defined as the region present at baseline but absent in the follow-up.

This resulted in a total of $4{,}896$ radiomic features ($204$ feature types $\times$ $4$ MRI modalities $\times$ $2$ timepoints $\times$ $3$ mask types), capturing information about tumor shape and texture.

Second, we engineered features to describe the overall tumor evolution. This included a \textit{centroid\_shift\_mm} feature, quantifying the spatial displacement of the tumor's center of mass between timepoints, and a \textit{time\_gap\_weeks} feature to account for the variable duration between the baseline and follow-up scans.

Third, we incorporated information based on the three primary tumor compartments: the Necrotic/Non-Enhancing core, the Enhancing Core, and the surrounding Edema. For each compartment, we calculated the baseline volume, follow-up volume, and the relative volume change. 

\subsubsection{Feature Fusion and Classification with CatBoost.}
\begin{figure}[!htb]
\begin{center}
\includegraphics[width=\columnwidth]{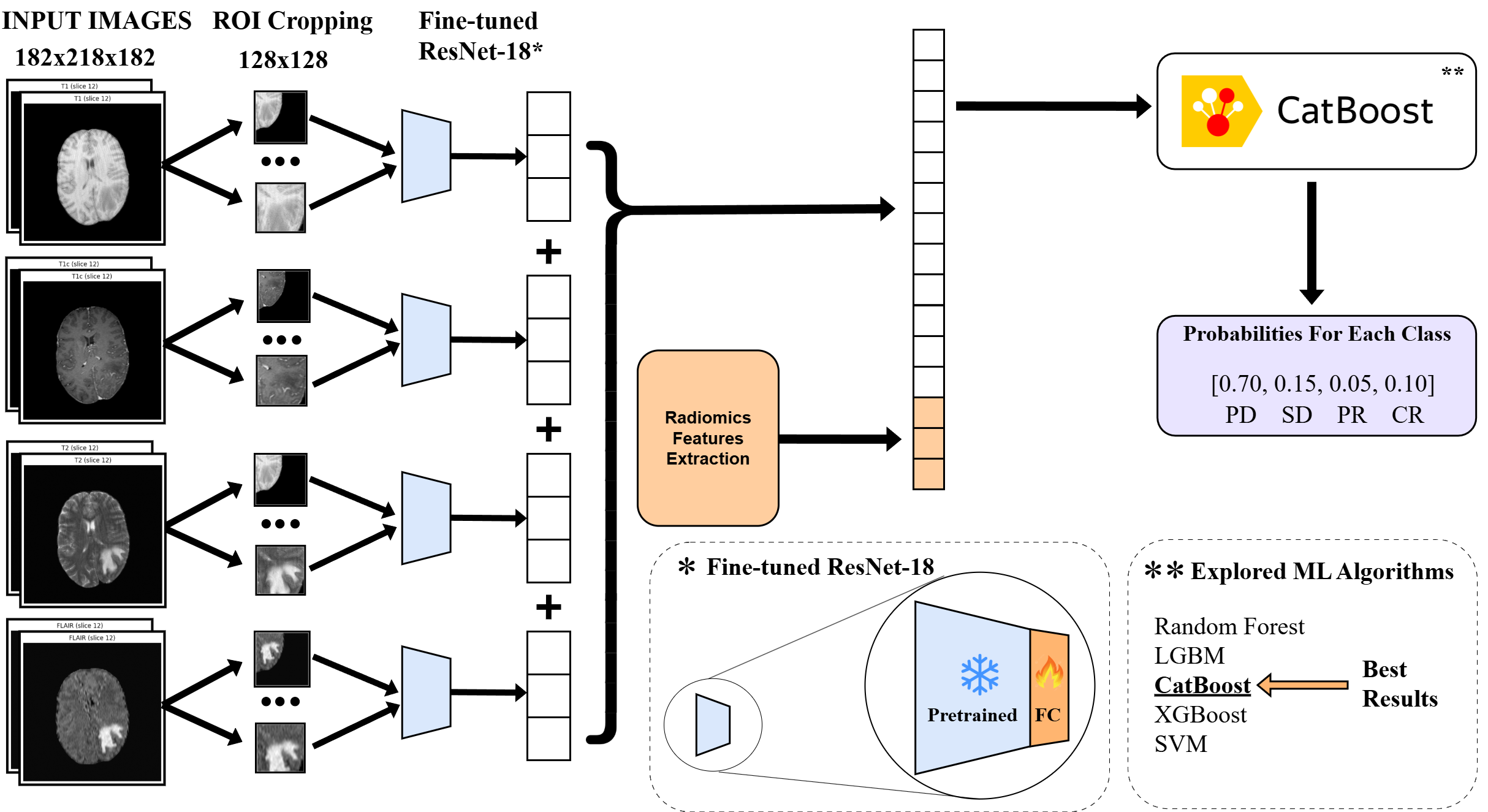}
\caption{Illustration of how The final feature vector was created and used as input to CatBoost.}
\label{Fig2}
\end{center}
\end{figure}

The final feature vector for each case was constructed by concatenating the deep features from the ResNet-18 encoders with the entire suite of 4,896+ handcrafted radiomic, volumetric, and engineered features (as shown in Figure~\ref{Fig2}).
For the classification task, we evaluated a range of powerful machine learning models, including XGBoost, Random Forest, LightGBM, and Support Vector Machines (SVM). Through our experimentation, the CatBoost model \cite{Prokhorenkova2018}, a gradient boosting framework that excels at handling heterogeneous features and categorical data, demonstrated superior performance. The trained CatBoost model outputs a probability vector for the four RANO response classes, from which the final prediction is derived.

\subsection{Experimental Setup and Evaluation}
\begin{figure*}[htb]
\centering
\subfigure[Label distribution in the training set across the four RANO classes.]{
    \includegraphics[width=0.45\textwidth]{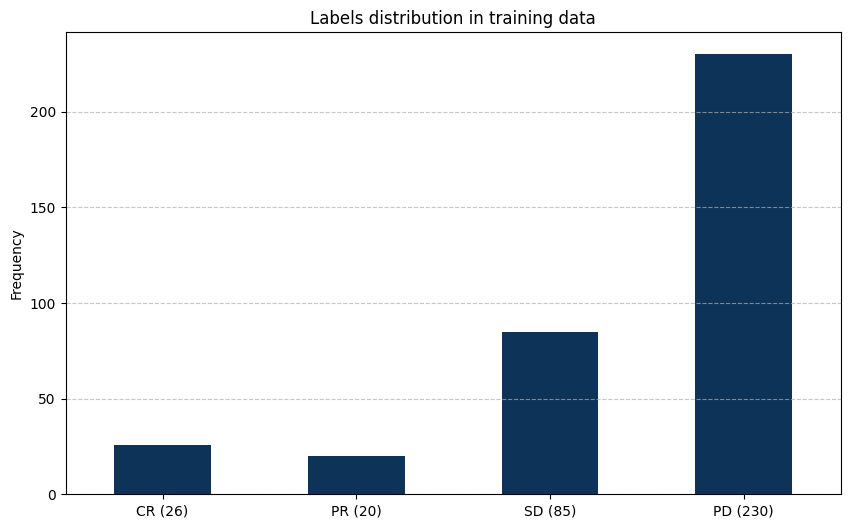}
    \label{Fig_t_dist}
}
\hfill
\subfigure[Label distribution per fold after patient-wise, stratified splitting.]{
    \includegraphics[width=0.45\textwidth]{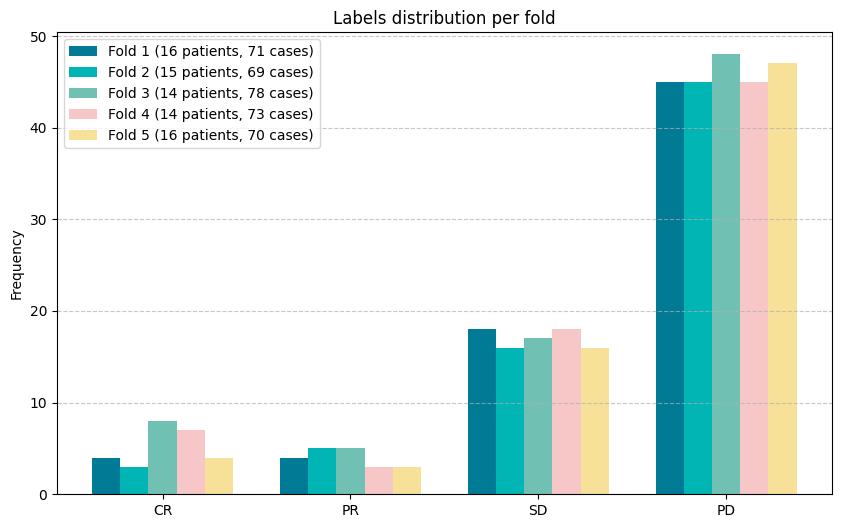}
    \label{Fig_folds}
}
\caption{Visualization of the dataset's label distribution and the stratified 5-fold patient-wise splitting strategy.}
\end{figure*}

To ensure a robust and unbiased evaluation of our model, all experiments were conducted using a 5-fold cross-validation scheme. Folds were split patient-wise, ensuring that all scans from a single patient were confined to a single fold, to prevent data leakage between training and validation sets. The splits were also stratified to maintain a similar distribution of the four RANO classes and to balance the total number of cases within each fold (Fig.~\ref{Fig_t_dist} and Fig.~\ref{Fig_folds}).

Model performance was assessed using three standard metrics. We used the Macro F1-score to evaluate the model's effectiveness across all classes, giving equal weight to each class regardless of its frequency. We also measured the area under the receiver operating characteristic curve (ROC AUC), averaged across classes, to assess the model's overall discriminative power. Finally, we report the overall Accuracy. The results presented in the following section are the mean and standard deviation of these metrics across all five folds.

\section{Results}
The performance of our proposed methodology was evaluated through a series of experiments designed to assess the contribution of each major component of our feature set. The results, averaged over the 5-fold patient-wise cross-validation, are presented in Table~\ref{tab:results}. We report the Macro F1-score, ROC AUC, and overall Accuracy to provide a comprehensive view of model performance.

\begin{table}[htbp]
\centering
\caption{Performance comparison of different feature sets for RANO response classification. Each experiment builds upon the previous ones, demonstrating the incremental value of each component. Results are shown as mean $\pm$ standard deviation across 5 cross-validation folds. The best performance for each metric is highlighted in bold.}\label{tab:results}
\begin{tabular}{|l|c|c|c|}
\hline
\bfseries Approach & \bfseries F1-score (Macro) & \bfseries ROC AUC & \bfseries Accuracy \\
\hline
Volumes only & 0.30 $\pm$ 0.02 & 0.59 $\pm$ 0.05 & 0.59 $\pm$ 0.05 \\
Radiomics from Raw dataset & 0.34 $\pm$ 0.09 & 0.70 $\pm$ 0.06 & 0.46 $\pm$ 0.06 \\
Growth/Shrinkage masks & 0.45 $\pm$ 0.08 & 0.73 $\pm$ 0.07 & 0.67 $\pm$ 0.07 \\
2D Radiomics + centroid shift & 0.44 $\pm$ 0.09 & 0.78 $\pm$ 0.06 & 0.68 $\pm$ 0.04 \\
CV (ResNet-18) & 0.43 $\pm$ 0.09 & 0.74 $\pm$ 0.07 & 0.64 $\pm$ 0.04 \\
Radiomics + CV (Final Model) & \bfseries 0.50 $\pm$ 0.08 & \bfseries 0.81 $\pm$ 0.08 & \bfseries 0.72 $\pm$ 0.05 \\
\hline
\end{tabular}
\end{table}

The experimental results clearly demonstrate the effectiveness of our hybrid approach. A baseline model using only simple volumetric features yielded limited predictive power, with a Macro F1-score of 0.30 and a ROC AUC of 0.59. The introduction of a comprehensive 3D radiomics feature set provided a notable increase in ROC AUC to 0.70, underscoring the value of textural and morphological information, although accuracy saw a decrease.

A significant performance gain was observed upon the inclusion of our specialized growth and shrinkage masks for radiomics calculation (Approach 3), which boosted the Macro F1-score to 0.45 and accuracy to 0.67. This highlights the importance of explicitly modeling the dynamic changes in the tumor between timepoints. The addition of 2D radiomics and the tumor centroid shift feature further improved the ROC AUC to 0.78.

Interestingly, the deep learning model (CV ResNet-18) used in isolation (Approach 5) achieved respectable performance, with a ROC AUC of 0.74, but was outperformed by our more advanced radiomics models. The true strength of our methodology is shown in our final model (Approach 6), which combines the ResNet-18 image representations with the full suite of radiomic and engineered features. This synergistic model achieved the highest scores across all metrics, with a Macro F1-score of 0.50, a ROC AUC of 0.81, and an accuracy of 0.72. This confirms our central hypothesis that integrating learned deep features with domain-specific handcrafted features provides the most robust and accurate framework for this complex classification task.

\section{Discussion}
In this study, we developed and validated a hybrid deep learning and radiomics framework for the automated classification of tumor response in glioblastoma patients, as part of the BraTS 2025 Challenge. Our primary finding is that a model integrating features from multiple sources—learned deep representations, extensive 3D and 2D radiomics, and domain-specific engineered features—yields superior performance compared to models relying on any single feature type alone. The final model's strong performance, achieving a mean ROC AUC of 0.81 and a Macro F1-score of 0.50, underscores the value of this synergistic approach.

The incremental performance gains demonstrated in our ablation study (Table~\ref{tab:results}) offer significant insights. The move from simple volumetric data to comprehensive radiomics, particularly those derived from our novel growth and shrinkage masks, provided one of the most substantial improvements. This highlights that capturing the texture and morphology of the evolving tumor regions is more informative than assessing volume alone. The results suggest that deep learning models and handcrafted features capture complementary information. While the ResNet-18 encoders learn abstract, hierarchical patterns from the raw image data, the handcrafted features ensure that clinically established concepts, such as volumetric changes and shifts in tumor location, are explicitly encoded and available to the classifier.

The strength of our work lies in the comprehensive nature of our feature engineering and the rigorous, patient-wise cross-validation that ensures the robustness of our findings. By creating separate features for stable, growing, and shrinking tumor regions, our model is designed to be sensitive to the complex and heterogeneous changes that characterize tumor response to therapy.

Nevertheless, we acknowledge several limitations in our current approach. A primary concern is the computational expense of our pipeline. The extensive preprocessing, including atlas registration and bias field correction, requires approximately five minutes for each case (baseline and follow-up scan pair), while the extraction of over 4,800 radiomic features adds another 40 seconds. This computational burden, while manageable for an offline challenge, could hinder rapid prototyping of new ideas and pose a challenge for eventual real-time clinical application.

Secondly, like many studies in medical imaging, our model's development is constrained by the size of the available annotated dataset. While the LUMIERE dataset \cite{Suter2022} is a high-quality resource, deep learning models in particular benefit from very large datasets. A larger and more diverse multi-centric cohort would be valuable for improving the model's generalizability and reducing the risk of overfitting, ensuring that it is robust to variations in imaging protocols and patient populations.

These limitations directly inform our directions for future work. The central avenue is the collaborative effort to build and share larger, more comprehensive annotated longitudinal datasets. For our model specifically, future work will involve pipeline optimization. This could include implementing feature selection algorithms to identify the most and least important radiomic features, thereby reducing dimensionality and computational overhead without sacrificing predictive power. Finally, the framework could be expanded to incorporate non-imaging data, such as patient demographics, or treatment specifics, to create a more holistic and potentially even more accurate predictive model.

\section{Conclusion}

In this work, we present a robust hybrid framework for predicting RANO-defined treatment response in glioma patients. Our approach fuses deep learning features extracted via fine-tuned ResNet-18 model with an extensive set of radiomic, volumetric, and engineered features that capture both spatial and temporal tumor characteristics. By integrating these complementary feature types, the model achieves accurate and objective response prediction, contributing to the advancement of automated neuro-oncological assessment.

Despite its promising performance, our method has some limitations. First, the reliance on pre-defined tumor masks may introduce variability due to segmentation quality. Second, the model's generalizability to multi-institutional datasets with heterogeneous imaging protocols remains to be validated.

Future work will focus on improving robustness through automated uncertainty estimation and testing across diverse clinical datasets. Additionally, exploring transformer-based architectures for spatiotemporal modeling may further enhance predictive performance.

\begin{credits}
\subsubsection{\ackname} The authors acknowledge that this research was funded by the Mohamed bin Zayed University of Artificial Intelligence (MBZUAI).

\subsubsection{\discintname}
The authors have no competing interests to declare that are relevant to the content of this article.
\end{credits}
%
%
%
%

\end{document}